\documentclass[10pt, a4paper]{article}
\usepackage{lrec}
\usepackage{multibib}
\newcites{languageresource}{Language Resources}
\usepackage{graphicx}
\usepackage{tabularx}
\usepackage{soul}

\usepackage{epstopdf}
\usepackage[latin1]{inputenc}

\usepackage{hyperref}
\usepackage{xstring}

\usepackage{todonotes}
\usepackage{booktabs}
\usepackage{mathptmx}

\title{Cross-lingual Candidate Search for Biomedical Concept Normalization}

\name{Roland Roller$^{1}$, Madeleine Kittner$^{2}$, Dirk Weissenborn$^{1}$, Ulf Leser$^{2}$}

\address{$^{1}$German Research Center for Artificial Intelligence (DFKI), Language Technology, Berlin, Germany \\ 
$^{2}$Humboldt Universit\"at zu Berlin, Knowledge management in Bioinformatics, Berlin, Germany \\
{\tt \{firstname.surname\}@dfki.de}, \tt \{surname\}@informatik.hu-berlin.de }
         
\abstract{
Biomedical concept normalization links concept mentions in texts to a semantically equivalent concept in a biomedical knowledge base. This task is challenging as concepts can have different expressions in natural languages, e.g. paraphrases, which are not necessarily all present in the knowledge base. Concept normalization of non-English biomedical text is even more challenging as non-English resources tend to be much smaller and contain less synonyms.
To overcome the limitations of non-English terminologies we propose a cross-lingual candidate search for concept normalization using a character-based neural translation model trained on a multilingual biomedical terminology. Our model is trained with Spanish, French, Dutch and German versions of UMLS.
The evaluation of our model is carried out on the French Quaero corpus, showing that it outperforms most teams of CLEF eHealth 2015 and 2016. Additionally, we compare performance to commercial translators 
on Spanish, French, Dutch and German versions of Mantra. Our model performs similarly well, but is free of charge and can be run locally. This is particularly important for clinical NLP applications as medical documents underlay strict privacy restrictions.
\newline }

\begin{document}

\maketitleabstract

\section{Introduction}
Concept normalization is the task of linking a text mention to a corresponding concept in a knowledge base (KB). This is useful to determine its distinct meaning and to include additional information linked through that knowledge base. The Unified Medical Language System (UMLS)~\cite{bodenreider2004unified} is a large biomedical knowledge base which unifies different terminologies and their concepts, also across languages.
Although UMLS includes synonyms and lexical variants for each concept, these are usually not exhaustive, since mentions in natural language can be expressed in many different ways which makes the task of concept normalization challenging. When dealing with non-English biomedical or clinical texts concept normalization becomes even more difficult as compared to English other languages are underrepresented in UMLS in terms of number of concepts or synonyms. Currently, UMLS\footnote{https://www.nlm.nih.gov/research/umls/knowledge\_sources /metathesaurus/release/statistics.html, accessed January 7th 2018} includes 25 different languages represented by 13,897,048 concept names (terms) which describe 3,640,132 individual concepts. The majority of concept names are English ($\approx70\%$). Concept names in other languages make out a much smaller part: for instance, Spanish $\approx10\%$, French $\approx3\%$, Dutch $\approx2\%$, and German $\approx2\%$.


To support non-English biomedical concept normalization two approaches can be observed: (i) translating terms or whole documents from the target language to English and search in the English knowledge base or (ii) translating relevant subsets of the English knowledge base to the target language in order to expand the target knowledge base. Both approaches have been used with varying success for different languages, for instance for French~\cite{Afzal:2015,Jiang:2015,Mulligen:2016} or Italian~\cite{chiaramello2016use}.


All of these studies apply commercial tools such as Google Translate\footnote{\url{https://www.translate.google.com}} or Bing Translator\footnote{\url{https://www.bing.com/translator}} for translation. 
Despite the good results, web-based translators can not be used when dealing with clinical documents. These data underly strict privacy restrictions and can not be shared online. Therefore, adaptable, local translation models are needed for NLP research in the biomedical and clinical domain.

In this work, we present a sequential cross-lingual candidate search for biomedical concept normalization. The central element of our approach is a neural translation model trained on UMLS for Spanish, French, Dutch and German. 
Evaluation on the French Quaero corpus shows that our approach outperforms most teams of CLEF eHealth 2015 and 2016. On Mantra we compare the performance of our translation model to commercial translators (Google, Bing) for Spanish, French, Dutch and German. Our model\footnote{The model is available here: \url{http://macss.dfki.de}.} performs similarly well, but can be run locally and is free of charge.

\section{Related work}
Concept normalization of non-English biomedical text has been the subject of several CLEF challenges. In CLEF eHealth 2015 Task 1b~\cite{Neveol:2015} and 2016 Task 2~\cite{Neveol:2016} named entity recognition and normalization was performed on the French Quaero corpus containing Medline titles and EMEA abstracts. Apart from other tasks teams were asked to perform concept normalization using gold standard annotations. 

The best performing team in 2015, team Erasmus~\cite{Afzal:2015} used a rule based dictionary lookup approach. Erasmus expanded the French version of UMLS by translating a potentially interesting subset of English UMLS to French using Google Translator and Bing Translate. Translations were only used when both translation systems returned the same result. Additionally, they applied several post-processing rules developed from the training data to remove false positives: preferring most frequently used concept IDs for certain terms or most frequent semantic type and concept ID pairs. 
The winning team in 2016, team SIBM~\cite{Cabot:2016} used the web-based service ECMT (Extracting Concepts with Multiple Terminologies) which performs bag of words concept matching at the sentence level. ECMT integrates up to 13 terminologies partially or totally translated into French. 
Compared to those teams we do not translate terminologies but terms and search in full non-English and English subsets of UMLS. Additionally, we use a similar disambiguation procedure as \cite{Afzal:2015}.

One team~\cite{Jiang:2015} in 2015 translated gold standard annotations to English using Google Translate and then applied MetaMap~\cite{Aronson:2010} for normalization. This approach only yielded moderate results which has similarly been shown for Italian~\cite{chiaramello2016use}. We apply a sequential search for candidate concepts. English UMLS is only used when the search in Non-English UMLS was not successful. A similar procedure is usually applied when initially annotating non-English corpora~\cite{Neveol:2014quaero,Kors:2015}.

Most teams in CLEF eHealth 2015 and 2016 rely on commercial online translation systems. 
Since the use of such tools is questionable and most probably forbidden when dealing with medical text local solutions are needed. Local machine translation models for the biomedical domain have been developed previously, for instance as part of the CLEF ER challenge 2013~\cite{rebholz2013entity} based on the parallel English, Spanish, French, Dutch and German Mantra corpus. Participating teams often used phrase-based statistical machine translation models~\cite{Attardi:2013,bodnari2013multilingual,hellrich2013julie}. However, to develop more sophisticated neural translation models large sets of parallel sentences are required, whereas the Mantra corpus is rather small. We developed a neural translation model on parallel language data of UMLS and FreeDict to be used for cross-lingual candidate search in concept normalization.

\section{A Neural Translation Model for concept normalization using UMLS}
The following section describes the biomedical translation model and the sequential procedure for candidate search during concept normalization.

\subsection{Translation Model}
The central element of our cross-lingual concept normalization solution is a character-based neural translation model~\cite{Lee:2016}. The model does not require any form of segmentation or tokenization at all. We chose this model because many translations of biomedical concepts can be resolved by small amends, due to the common origin of many words, that can be captured by such a system. At the same time it has enough modeling capacity to learn translations rules that cannot be captured by simple surface-form rules.

\paragraph{Model}
The model embeds lower-cased characters of the source phrase into a $256$-dimensional space. The embedded character sequence is processed by a convolution layer with $N = 16 + 32 + 64 + 64 + 128 + 128 + 256 = 688$ filters of varying width resulting in $688$-dimensional states for each character position. 
These are $\max$-pooled over time within fixed, successive intervals of $k = 5$. This effectively reduces the number of source states by a factor of $5$. To allow for additional interaction between the pooled states, they are further transformed by a 2-layer highway network. 
Finally, the transformed states are processed by a bidirectional recurrent neural network, in particular a bidirectional GRU~\cite{chung:2014}, to produce final encoder states. A 2-layer recurrent neural network with attention~\cite{Bahdanau:2015} on the encoder states subsequently produces the translation character by character. For more in-depth, technical details we refer the reader to~\cite{Lee:2016}.  The model is trained on mini-batches comprising $32$ source phrases with their respective translations, using ADAM~\cite{Kingma:2015} as optimizer. The initial learning rate is set to $10^{-3}$ which is halved whenever performance on the development set drops.

\paragraph{Dataset}
We train our system on a subset of UMLS concept translations combined with various English to target language dictionaries taken from the FreeDict project~\footnote{\url{http://freedict.org/en/}}.
There might be multiple target language translations for each English phrase, thus we employ a deterministic decoder. The model is trained to minimize the
perplexity only on the transformation that is most likely under the current model, i.e., the transformation with the least loss. 

\subsection{Concept Normalization}
We approach concept normalization in two steps. First a candidate search is carried out while terms are sequentially looked up in non-English and English versions of UMLS. This first step aims at achieving a high recall. Then a disambiguation step is applied in order to reduce the number of candidates while keeping the precision high. In the following details on both steps are described.

\begin{figure}[bht!]
\centering
  \includegraphics[width=0.34\textwidth]{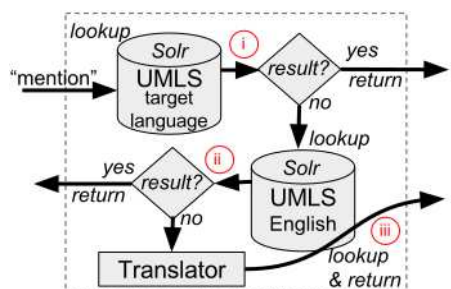}
  \caption{\label{tab:fig1} Sequential Candidate Search using mono-lingual (target lang.) and cross-lingual (English) UMLS subsets.}
    \vspace{-0.2cm}
\end{figure}

\paragraph{Candidate Search}
Concept terms of English, Spanish, French, Dutch and German versions of UMLS AB2017 are indexed and searched using Apache Solr 6.5.0\footnote{\url{http://lucene.apache.org/solr/}}. A dictionary lookup always applies exact matching. If this first search does not return any results fuzzy matching is applied. Fuzzy matching uses a Levensthein edit distance of one per token for tokens larger than 4 characters otherwise the edit distance is set to zero.

As shown in Figure \ref{tab:fig1}, we apply a sequential procedure for candidate search. Subsequently, the following searches are performed: (i) a mono-lingual candidate search looking up terms only in the respective non-English UMLS, (ii) a simple cross-lingual candidate search looking up the original term in English UMLS, and (iii) a cross-lingual candidate search in English UMLS while terms are first translated to English using our biomedical translation model. Once matching candidates are found the sequence is stopped. In this work we study how the addition of each search level improves concept normalization. We compare candidate searches up to level (i) mono-lingual (ML), level (ii) simple cross-lingual (CL), and level (iii) cross-lingual including translation of terms (BTM). 


\paragraph{Disambiguation}

Our sequential candidate search may return a list of candidate concepts. This list is filtered by the following steps: (1) filter for known UMLS semantic groups or types, (2) prefer concepts with UMLS preferred labels, (3) filter by using densest-subgraph disambiguation \cite{Moro14entitylinking,Weissenborn2016}, and (4) choose the smallest UMLS concept ID. 
Steps 1, 2, and 4 were similarly applied by \cite{Afzal:2015}. 

\section{Evaluation}
The performance of our sequential candidate search including our biomedical translation model is evaluated on two corpora: Quaero~\cite{Neveol:2014quaero} and Mantra~\cite{Kors:2015}. In the following the corpora and evaluation procedures are explained and results are presented.

\subsection{Evaluation Corpora}\label{sec:evalCorpora}

\paragraph{Quaero} The Quaero corpus contains French Medline titles and EMEA abstracts and has been used for named entity recognition and normalization tasks in CLEF eHealth 2015 Task 1b~\cite{Neveol:2015} and 2016 Task 2~\cite{Neveol:2016}. We compare our approach to results of teams performing best in the entity normalization task, \newcite{Afzal:2015} and \newcite{Cabot:2016}. For that purpose we extracted gold standard annotations from the test corpora in 2015 and 2016. 
Note, the current (2016) version of Quaero contains a training, development and test set. The current development set is the test set of 2015.


\paragraph{Mantra} The Mantra corpus contains Medline titles, EMEA abstracts and EPO patents for several languages including bi-lingual aligned sentences. For evaluation of our system we extracted gold standard annotations of Mantra Medline titles in Spanish, French, Dutch, and German. 
Note, that the Mantra Medline corpus is much smaller than the Quaero corpus. For evaluation we compare perfomance of our system to performance of commercial translators. Hereby, we apply the same sequential candidate search while instead of our biomedical translation model we use translations obtained manually from Google Translate and Bing Translator 
Similar to~\newcite{Afzal:2015}, only the first translation of each service was selected and used only if both systems returned the same translation.


\subsection{Evaluation Results for Quaero}

We evaluate performance of our proposed method against best performing systems in CLEF eHealth challenges 2016 and 2015. In 2016, team SIBM~\cite{Cabot:2016} performed best on the task of gold standard entity normalization, see Table~\ref{evaluation_results_quaero_test} for their results. 
In 2015, team Erasmus~\cite{Afzal:2015} performed far better. They achieved an F1-score of 0.872 for EMEA and 0.671 for Medline, see Table~\ref{evaluation_results_quaero_dev}. Moreover, the system was able to achieve a precision of 1 for EMEA. 

\begin{table}[ht!]
  \centering
  \begin{tabular}{ @{\hskip 2mm}l@{\hskip 2mm}  @{\hskip 3mm}c@{\hskip 2mm}c@{\hskip 2mm}c@{\hskip 2mm}  @{\hskip 3mm}c@{\hskip 2mm}c@{\hskip 2mm}c@{\hskip 2mm} }
    \hline
    & \multicolumn{3}{ c@{\hskip 1mm}}{\textbf{Medline}} & \multicolumn{3}{ c }{\textbf{EMEA}} \\
	Method & P & R & F1 & P & R & F1 \\
	\hline
 	ML & \textbf{0.800} & 0.594 & 0.682 & \textbf{0.822} & 0.552 & 0.661 \\ 
 	CL & 0.786 & 0.620 & 0.693 & 0.808 & 0.676 & 0.736 \\ 
 	BTM & 0.771 & \textbf{0.663} & \textbf{0.713} & 0.781 & \textbf{0.692} & \textbf{0.734} \\ 
	\hline
	SIBM & 0.594&0.515&0.552 & 0.604&0.463&0.524 \\ 
    \hline
    \end{tabular}
        \caption{Evaluation of mono- and cross-lingual candidate search for concept normalization on Quaero Corpus of CLEF eHealth challenge 2016 Task 2. We compare against SIBM~\protect\cite{Cabot:2016}, the winning system of the challenge. Methods presented include mono-lingual (ML) and cross-lingual (CL) candidate search, and cross-lingual candidate search including translation of concept terms using our biomedical translation model (BTM). BTM outperforms SIBM on both, Medline and EMEA.}
\label{evaluation_results_quaero_test}
\end{table}

\begin{table}[ht!]
  \centering
  \begin{tabular}{ @{\hskip 2mm}l@{\hskip 2mm}  @{\hskip 3mm}c@{\hskip 2mm}c@{\hskip 2mm}c@{\hskip 2mm}  @{\hskip 3mm}c@{\hskip 2mm}c@{\hskip 2mm}c@{\hskip 2mm} }
    \hline
    & \multicolumn{3}{ c@{\hskip 1mm}}{\textbf{Medline}} & \multicolumn{3}{ c }{\textbf{EMEA}} \\
	Method & P & R & F1 & P & R & F1 \\
	\hline
    ML & 0.831 & 0.575 & 0.680 & 0.911 & 0.632 & 0.746 \\ 
    CL & \textbf{0.834} & 0.611 & 0.705 & 0.919 & 0.764 & 0.834 \\ 
    BTM & 0.831 & \textbf{0.661} & \textbf{0.736} & 0.909 & 0.772 & 0.835 \\ 
    \hline
    Erasmus & 0.805 & 0.575 & 0.671 & \textbf{1.000} & \textbf{0.774} & \textbf{0.872} \\
    \hline
   \end{tabular}
    \caption{Evaluation of mono- and cross-lingual candidate search for concept normalization on Quaero Corpus of CLEF eHealth challenge 2015 Task 1b. We compare against Erasmus~\protect\cite{Afzal:2015}, the winning system of the challenge. Methods presented include mono-lingual (ML) and cross-lingual (CL) candidate search, and translation of concept terms using our biomedical translation model (BTM). BTM outperforms Erasmus on Medline but not on EMEA.}
  \label{evaluation_results_quaero_dev}
\end{table}

\begin{table*}[ht!]
  \centering
  \begin{tabular}{ @{\hskip 3mm}l@{\hskip 3mm}  @{\hskip 4mm}c@{\hskip 3mm}c@{\hskip 3mm}c@{\hskip 3mm}  @{\hskip 4mm}c@{\hskip 3mm}c@{\hskip 3mm}c@{\hskip 3mm}  @{\hskip 4mm}c@{\hskip 3mm}c@{\hskip 3mm}c@{\hskip 3mm}  @{\hskip 4mm}c@{\hskip 3mm}c@{\hskip 3mm}c@{\hskip 3mm} }
    \hline
    & \multicolumn{3}{ c }{\textbf{SPA}} & \multicolumn{3}{ c }{\textbf{FRE}} & \multicolumn{3}{ c }{\textbf{DUT}} & \multicolumn{3}{ c }{\textbf{GER}} \\
    
    Method & P & R & F1 & P & R & F1 & P & R & F1 & P & R & F1 \\
    \hline
    ML & 
    \textbf{0.799} & 0.561 & 0.659 &  
    \textbf{0.814} & 0.469 & 0.595 & 
    \textbf{0.800} & 0.357 & 0.494 & 
    \textbf{0.833} & 0.493 & 0.620 \\ 

    CL & 
    0.788 & 0.583 & 0.670 &  
    0.795 & 0.502 & 0.615 & 
    0.769 & 0.424 & 0.546 & 
    0.817 & 0.530 & 0.643 \\ 
    
    BTM & 
    0.781 & \textbf{0.619} & \textbf{0.691} &  
    0.780 & 0.593 & 0.674 & 
    0.725 & 0.533 & 0.614 & 
    0.771 & 0.582 & 0.663 \\ 
    \hline
    GB & 
    0.790 & 0.607 & 0.687 & 
    0.794 & \textbf{0.604} & \textbf{0.686} & 
    0.767 & \textbf{0.560} & \textbf{0.648} & 
    0.804 & \textbf{0.588} & \textbf{0.679} \\ 
    \hline
     
    \end{tabular}
 
\caption{Evaluation on Medline titles of Mantra. We compare performance to commercial translation tools (GB) for Spanish, French, Dutch, and German. Methods presented include mono-lingual and cross-lingual candidate search, as well as our and biomedical translation model (BTM) or Google Translate and Bing Translator (GB).}
\label{evaluation_results}
\vspace{-0.2cm}
\end{table*}

Cross-lingual candidate search including our biomedical translation model outperforms results for entity normalization of previous teams in three out of four data sets. In 2016, mono-lingual candidate search reaches highest precision for Medline and EMEA and already outperforms SIBM, see Table~\ref{evaluation_results_quaero_test}. Why the system of SIBM yields such poor results is not clear. One reason could be that their integrated and translated terminologies have a lower coverage in terms of concepts as full UMLS. Although most terminologies they integrate are part of UMLS. For ranking extracted candidates they use a classification based on most relevant term-semantic type relations which might not be as sufficient as our disambiguation procedure. Extending the search space by using the English UMLS subset in addition (CL) leads to further improvements in terms of recall and F1. Finally, including our translation system, in combination with a cross-lingual search (BTM) leads to the highest recall and the highest F1-Score. 

On the dataset of 2015, we see the same pattern. Cross-lingual search improves the performance of the system in terms of recall and F1 in comparison to mono-lingual search and BTM outperforms CL. Moreover, CL and BTM both outperform Erasmus on the Medline dataset. On EMEA instead, our method cannot reach the performance of Erasmus. While recall of BTM and Erasmus are similar, the precision of our system is approximately 10\% lower. A reasonable explanation for that might be the fact that Erasmus applied various pre- and post-processing steps optimized for the Quaero copus. In contrast, our method is fully generic. Using a different and more corpus specific disambiguation might improve results for our system as well.

\subsection{Evaluation Results for Mantra}
We compare performance of our sequential mono- and cross-lingual candidate search for different languages, Spanish, French, Dutch, and German. Additionally, we directly compare performance of our biomedical translation model to commercial translators. The same sequential procedure is applied while instead of translating terms using our translation model we use translations from Google Translate and Bing Translator. Details on how translations with Google Translate and Bing Translator were obtained are described in Section \ref{sec:evalCorpora}.

Results presented in Table \ref{evaluation_results} show the same pattern for all languages as previous results for French on Quaero: Cross-lingual candidate search always outperforms mono-lingual candidate search and the integration of the translator outperforms CL and ML. The main reason for that is the boost of recall which leads, in combination with a good disambiguation, to a high precision and thus to an improved F1. 

Cross-lingual candidate search including our biomedical translation model (BTM) outperforms Google Translate and Bing Translator (GB) for Spanish in recall and F1. Overall, precision and recall are very similar for both systems and all languages. Differences in precision are very small for Spanish and French (0.01) and only slightly higher for Dutch and German (0.03-0.04). Differences in recall are in the same range: 0.02 for Spanish (while BTM outperforms GB), 0.01 for French, 0.03 for Dutch, and no significant difference for German.

We also compared performance of BTM and GB between languages. Although differences are small, in terms of recall and F1 the performance of both systems decreases in the following order SPA $>$ FRE $>$ GER $>$ DUT. Interestingly this order correlates well with the number of concepts for each language present in UMLS. In terms of precision, BTM shows the same order, while commercial translators yield best results for German: GER $>$ FRE $>$ SPA $>$ DUT.

\section{Conclusions}
In this work we present a character-based neural translation model trained on the multi-lingual terminology UMLS for Spanish, French, Dutch, and German. The model is integrated into a sequential candidate search for concept normalization. Evaluation on two different corpora shows that our proposed method significantly improves biomedical concept normalization for non-English texts.

We propose a sequential procedure for candidate search. Subsequently, terms are searched (i) in the relevant Non-English version of UMLS (ML), (ii) in English UMLS without translating the search term (CL), and (iii) translating the term using our biomedical translation model before searching in English UMLS (BTM). Once a matching concept is found the sequence is stopped. We chose this sequential procedure as ML and CL tend to result in a smaller number of false positives (data not shown here). 

In all evaluations we found that already a simple cross-lingual candidate search (CL) (without translation) improves recall significantly while at the same time the loss in precision is small. This might be explained by the fact that many biomedical terms are very similar across different languages because they originated from common Greek or Latin words. Therefore, already a fuzzy  search is able to detect the right concept for many terms.

We evaluated our system on previous concept normalization tasks using the French Quaero corpus. The sequential candidate search including our biomedical translation model outperformed the winning teams in 3 out of 4 data sets. In this study we focused on a novel approach for candidate search using established methods for disambiguation. Using corpus-specific disambiguation procedures might even further improve precision of our method.

Compared to commercial translation systems our biomedical translation model yields comparable results for French, Dutch, and German, and slightly outperforms on recall and F1 for Spanish. While commercial services require Internet access and are charged although with a low price
, our translation model is open-source and free of charge. Additionally, it can be run locally, and therefore be used for processing patient related clinical texts. Such data underly strict data privacy restrictions and are not allowed to be processed using online services. We assume integrating our translation model into NLP pipelines will improve results for non-English biomedical information extraction tasks.


\section*{Acknowledgements}
This research was supported by the German Federal Ministry of Economics and Energy (BMWi) through the project MACSS (01MD16011F) and the German Federal Ministry of Education and Research (BMBF) through the project PERSONS (031L0030B).

\section{Bibliographical References}
\label{main:ref}

\bibliographystyle{lrec}
\bibliography{xample}

\begin{thebibliography}{}

\bibitem[\protect\citename{Afzal \bgroup et al.\egroup }2015]{Afzal:2015}
Afzal, Z., Akhondi, S.~A., van Haagen, H., van Mulligen, E.~M., and Kors, J.~A.
\newblock (2015).
\newblock {Biomedical Concept Recognition in French Text Using Automatic
  Translation of English Terms.}
\newblock In {\em CLEF (Working Notes)}.

\bibitem[\protect\citename{Aronson and Lang}2010]{Aronson:2010}
Aronson, A.~R. and Lang, F.-M.
\newblock (2010).
\newblock {An overview of MetaMap: historical perspective and recent advances}.
\newblock {\em Journal of the American Medical Informatics Association},
  17(3):229--236.

\bibitem[\protect\citename{Attardi \bgroup et al.\egroup }2013]{Attardi:2013}
Attardi, G., Buzzelli, A., and Sartiano, D.
\newblock (2013).
\newblock {Machine Translation for Entity Recognition across Languages in
  Biomedical Documents.}
\newblock In {\em CLEF (Working Notes)}. Citeseer.

\bibitem[\protect\citename{Bahdanau \bgroup et al.\egroup }2015]{Bahdanau:2015}
Bahdanau, D., Cho, K., and Bengio, Y.
\newblock (2015).
\newblock {Neural Machine Translation By Jointly Learning To Align and
  Translate}.
\newblock {\em ICLR}.

\bibitem[\protect\citename{Bodenreider}2004]{bodenreider2004unified}
Bodenreider, O.
\newblock (2004).
\newblock {The unified medical language system (UMLS): integrating biomedical
  terminology}.
\newblock {\em Nucleic acids research}, 32(suppl\_1):D267--D270.

\bibitem[\protect\citename{Bodnari \bgroup et al.\egroup
  }2013]{bodnari2013multilingual}
Bodnari, A., N{\'e}v{\'e}ol, A., Uzuner, {\"O}., Zweigenbaum, P., and
  Szolovits, P.
\newblock (2013).
\newblock {Multilingual Named-Entity Recognition from Parallel Corpora.}
\newblock In {\em CLEF (Working Notes)}. Citeseer.

\bibitem[\protect\citename{Cabot \bgroup et al.\egroup }2016]{Cabot:2016}
Cabot, C., Soualmia, L.~F., Dahamna, B., and Darmoni, S.~J.
\newblock (2016).
\newblock {{SIBM} at {CLEF} eHealth Evaluation Lab 2016: Extracting Concepts in
  French Medical Texts with {ECMT} and {CIMIND}}.
\newblock In {\em Working Notes of {CLEF} 2016 - Conference and Labs of the
  Evaluation forum, {\'{E}}vora, Portugal, 5-8 September, 2016.}, pages 47--60.

\bibitem[\protect\citename{Chiaramello \bgroup et al.\egroup
  }2016]{chiaramello2016use}
Chiaramello, E., Pinciroli, F., Bonalumi, A., Caroli, A., and Tognola, G.
\newblock (2016).
\newblock {Use of "off-the-shelf" information extraction algorithms in clinical
  informatics: A feasibility study of MetaMap annotation of Italian medical
  notes}.
\newblock {\em Journal of Biomedical Informatics}, 63:22--32.

\bibitem[\protect\citename{Chung \bgroup et al.\egroup }2014]{chung:2014}
Chung, J., Gulcehre, C., Cho, K., and Bengio, Y.
\newblock (2014).
\newblock Empirical evaluation of gated recurrent neural networks on sequence
  modeling.
\newblock {\em arXiv preprint arXiv:1412.3555}.

\bibitem[\protect\citename{Hellrich and Hahn}2013]{hellrich2013julie}
Hellrich, J. and Hahn, U.
\newblock (2013).
\newblock {The JULIE LAB MANTRA System for the CLEF-ER 2013 Challenge.}
\newblock In {\em CLEF (Working Notes)}.

\bibitem[\protect\citename{Jiang \bgroup et al.\egroup }2015]{Jiang:2015}
Jiang, J., Guan, Y., and Zhao, C.
\newblock (2015).
\newblock {WI-ENRE in CLEF eHealth Evaluation Lab 2015: Clinical Named Entity
  Recognition Based on CRF.}
\newblock In {\em CLEF (Working Notes)}.

\bibitem[\protect\citename{Kingma and Ba}2015]{Kingma:2015}
Kingma, D. and Ba, J.
\newblock (2015).
\newblock {Adam: A Method for Stochastic Optimization}.
\newblock {\em ICLR}.

\bibitem[\protect\citename{Kors \bgroup et al.\egroup }2015]{Kors:2015}
Kors, J.~A., Clematide, S., Akhondi, S.~A., van Mulligen, E.~M., and
  Rebholz-Schuhmann, D.
\newblock (2015).
\newblock {A multilingual gold-standard corpus for biomedical concept
  recognition: the Mantra GSC}.
\newblock {\em Journal of the American Medical Informatics Association},
  22(5):948.

\bibitem[\protect\citename{Lee \bgroup et al.\egroup }2016]{Lee:2016}
Lee, J., Cho, K., and Hofmann, T.
\newblock (2016).
\newblock {Fully Character-Level Neural Machine Translation without Explicit
  Segmentation}.
\newblock {\em CoRR}, abs/1610.03017.

\bibitem[\protect\citename{Moro \bgroup et al.\egroup
  }2014]{Moro14entitylinking}
Moro, A., Raganato, R., and Navigli, R.
\newblock (2014).
\newblock {Entity Linking meets Word Sense Disambiguation: A Unified Approach}.
\newblock {\em Transactions of the Association for Computational Linguistics}.

\bibitem[\protect\citename{N{\'e}v{\'e}ol \bgroup et al.\egroup
  }2014]{Neveol:2014quaero}
N{\'e}v{\'e}ol, A., Grouin, C., Leixa, J., Rosset, S., and Zweigenbaum, P.
\newblock (2014).
\newblock {The {QUAERO} {French} Medical Corpus: A Ressource for Medical Entity
  Recognition and Normalization}.
\newblock In {\em Proc of BioTextMining Work}, pages 24--30.

\bibitem[\protect\citename{N{\'e}v{\'e}ol \bgroup et al.\egroup
  }2015]{Neveol:2015}
N{\'e}v{\'e}ol, A., Grouin, C., Tannier, X., Hamon, T., Kelly, L., Goeuriot,
  L., and Zweigenbaum, P.
\newblock (2015).
\newblock {CLEF eHealth Evaluation Lab 2015 Task 1b: Clinical Named Entity
  Recognition.}
\newblock In {\em CLEF (Working Notes)}.

\bibitem[\protect\citename{N{\'e}v{\'e}ol \bgroup et al.\egroup
  }2016]{Neveol:2016}
N{\'e}v{\'e}ol, A., Goeuriot, L., Kelly, L., Cohen, K., Grouin, C., Hamon, T.,
  Lavergne, T., Rey, G., Robert, A., Tannier, X., et~al.
\newblock (2016).
\newblock {Clinical information extraction at the CLEF eHealth evaluation lab
  2016}.
\newblock In {\em Proceedings of CLEF 2016 Evaluation Labs and Workshop: Online
  Working Notes. CEUR-WS (September 2016)}.

\bibitem[\protect\citename{Rebholz-Schuhmann \bgroup et al.\egroup
  }2013]{rebholz2013entity}
Rebholz-Schuhmann, D., Clematide, S., Rinaldi, F., Kafkas, S., van Mulligen,
  E.~M., Bui, C., Hellrich, J., Lewin, I., Milward, D., Poprat, M., et~al.
\newblock (2013).
\newblock {Entity recognition in parallel multi-lingual biomedical corpora: The
  CLEF-ER laboratory overview}.
\newblock In {\em International Conference of the Cross-Language Evaluation
  Forum for European Languages}, pages 353--367. Springer.

\bibitem[\protect\citename{Van~Mulligen \bgroup et al.\egroup
  }2016]{Mulligen:2016}
Van~Mulligen, E., Afzal, Z., Akhondi, S.~A., Vo, D., and Kors, J.~A.
\newblock (2016).
\newblock {Erasmus MC at CLEF eHealth 2016: Concept recognition and coding in
  French texts}.
\newblock CLEF.

\bibitem[\protect\citename{Weissenborn \bgroup et al.\egroup
  }2016]{Weissenborn2016}
Weissenborn, D., Roller, R., Xu, F., Uszkoreit, H., and Perez, E.~G.
\newblock (2016).
\newblock {A Light-weight {\&} Robust System for Clinical Concept
  Disambiguation}.
\newblock In {\em Proceedings of the 7th International Symposium on Semantic
  Mining in Biomedicine, {SMBM} 2016, Potsdam, Germany, August 4-5, 2016.},
  pages 85--89.

\end{thebibliography}

\end{document}